\documentclass[runningheads]{llncs}
\usepackage{graphicx}
\usepackage{subfig}
\begin{document}
\title{Deformable Convolutional LSTM for Human Body Emotion Recognition}
%
%
\author{Peyman Tahghighi\inst{1} \and
Abbas Koochari \inst{2} \and
Masoume Jalali \inst{2}}
\authorrunning{P.Tahghighi et al.}
%
\institute{School of Electrical and Computer Engineering\\
University of Tehran,
Tehran, Iran \\
\email{Peyman.Tahghighi@ut.ac.ir}\\
\and
School of Mechanics, Electrical Power and Computer \\
Islamic Azad University, Science and Research Branch \\
Tehran, Iran}

\maketitle 
\begin{abstract}
People represent their emotions in a myriad of ways. Among the most important ones is whole body expressions which have many applications in different fields such as human-computer interaction (HCI). One of the most important challenges in human emotion recognition is that people express the same feeling in various ways using their face and their body. Recently many methods have tried to overcome these challenges using Deep Neural Networks (DNNs). However, most of these methods were based on images or on facial expressions only and did not consider deformation that may happen in the images such as scaling and rotation which can adversely
affect the recognition accuracy. In this work, motivated by recent researches on deformable convolutions, we incorporate the deformable behavior into the core of convolutional long short-term memory (ConvLSTM) to improve robustness to these deformations in the image and, consequently, improve its accuracy on
the emotion recognition task from videos of arbitrary length. We did experiments on the GEMEP dataset and achieved state-of-the-art accuracy of 98.8$\%$ on the task of whole human body emotion recognition on the validation set.

\end{abstract}
\section{Introduction}
Understanding and interpreting human emotions from videos have many different applications especially in the field of Human Computer/Robotic Interaction (HCI/HRI) \cite{16}. Humans can represent different emotions by their behavior using verbal and non-verbal signs during a conversation. Generally, human communication can be classified as verbal and nonverbal. The verbal communication includes voice and its tune, while nonverbal includes body movement, facial expression and gestures \cite{1,2}.

\par

So far, many researchers have tried to analyze and classify emotion from a single image or a stream of frames. For instance, Jain \emph{et al.} \cite{5}, used images of the face and a deep convolutional neural network (CNN) based on ResNet \cite{6} for classification of emotions based on human faces, while \cite{7} used a similar idea to classify whole body emotion. Nevertheless, both of these methods were based on fixed images, while we are trying to classify emotion from videos. Moreover, Jeong \emph{et al.} \cite{8} used 3D CNNs on a stream of facial videos for classification, but their method was limited to facial videos. Considering whole human body emotion recognition, Ahmed \emph{et al.} \cite{14} used features extracted from human body movement to classify different emotions, while Santhoshkumar \emph{et al.} \cite{15} used features extracted from two consecutive frames using their difference. Some other methods relied on multi modalities such as video and audio. For instance, Chen \emph{et al.} \cite{9} leveraged both audio and video in a multiple feature fusion method to classify videos. However, in this paper, we try to use video frames only and the whole human body to classify emotions.

\par

Motivated by the recent research on deformable 2D convolutions \cite{10,11} which showed its effectiveness in classification and object recognition, in this paper, we tried to incorporate deformable convolutions into the core of ConvLSTMs \cite{13} in order to improve its flexibility to detect and extract features from a given frame. That is to say, since the offsets in the deformable convolutional layer allow free deformation of the sampling grid (receptive field) in comparison to fix grid in regular convolution layers, they can perform better at capturing and focusing on the salient part of frames. Furthermore, we combined deformable ConvLSTMs with 3D convolutions to extract both long-term and short-term spatio-temporal features from videos. Afterward, we used a shallow 2D CNN architecture to extract features from each 2D spatio-temporal feature map individually and finally, we used feature fusion for final classification. We experimented with GEMEP \cite{4} dataset which contains 145 videos representing 17 different emotions and achieved state-of-the-art accuracy of 98.8$\%$ on the validation set.
\section{Method}
\subsection{Input preprocessing}
Different individuals may represent different emotions at various speeds. Hence, videos in our dataset contain an arbitrary number of frames and we had to set all videos to a fixed number of frames. One of the ways was to split each video into an array of videos with fixed frames, but one clip sometimes cannot represent the whole emotion. Consequently, we used the idea of Uniform sampling with temporal jitter \cite{12} to fix the length of all videos to 32 frames. If the length of one video was less than 32, we repeated the last frame. Moreover, we reduced the original frame size of videos from $720 \times 576$ to $112 \times 112$ to reduce the parameter size of our network.

\subsection{Network architecture}
\begin{figure}[h]
\centering
\includegraphics[clip,trim=4.8cm 19.8cm 1cm 0.45cm]{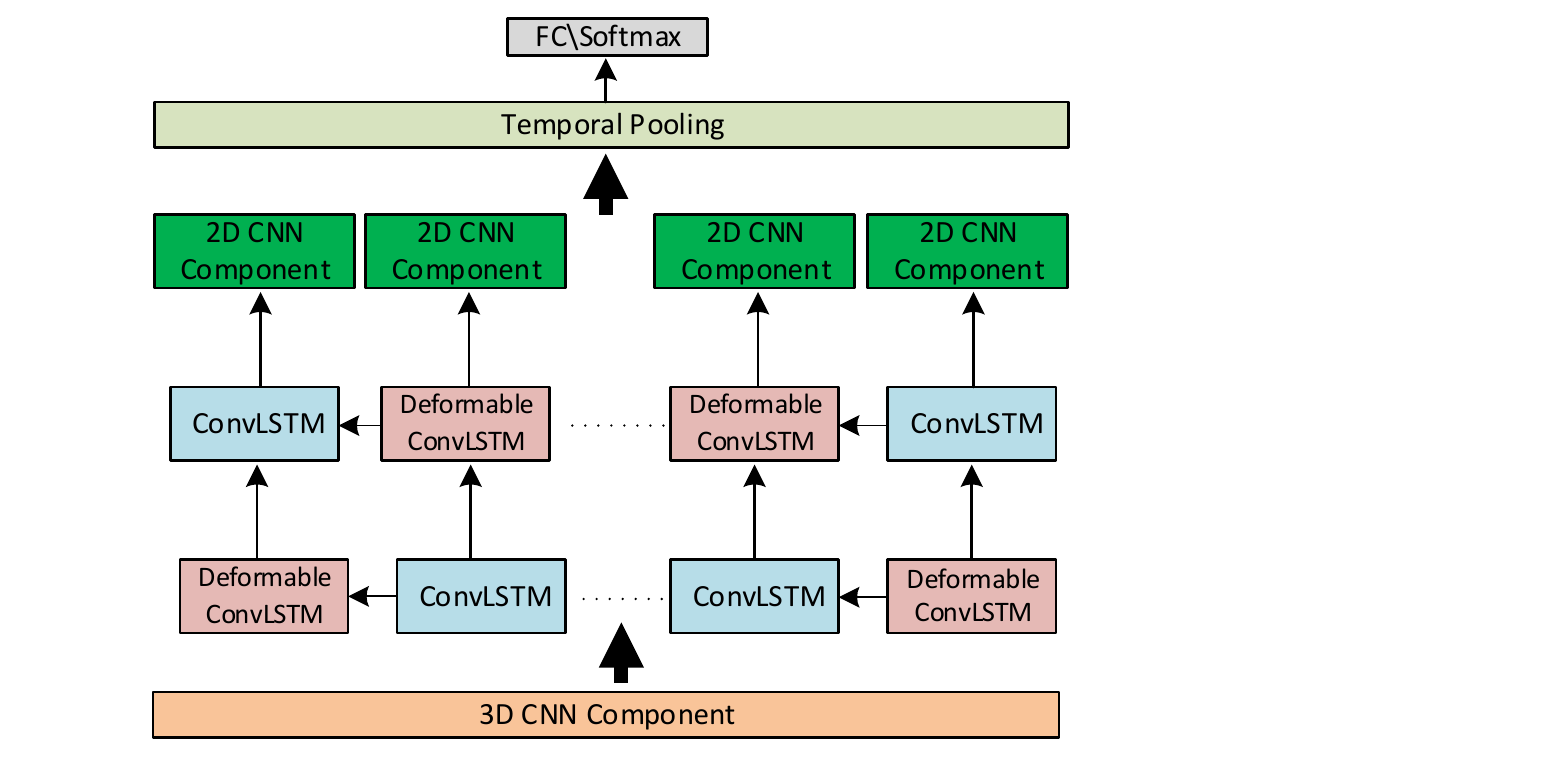}
\caption{Network architecture for emotion classification.}
\label{fig1}
\end{figure}

As can be seen in Figure \ref{fig1}, our network is consists of three major components which in this section we elaborate on each of them thoroughly.
\subsubsection{3D CNN component}
Inspired by \cite{12}, since 3D convolutions can perform better at learning local and short-term spatio-temporal features, in the first layer we used a 3D CNN component. As can be seen in Figure \ref{fig2a}, we only used two max pooling operations which only one of them shrinks the video length. Consequently, this component only focuses on the short-term spatio-temporal features and we leave the learning of long-term spatio-temporal features to the deformable convolutional LSTMs.

\subsubsection{Deformable convolutional LSTMs}
Recurrent Neural Networks (RNNs) and LSTMs are more suitable for learning long-term spatio-temporal features \cite{12}. Nevertheless, ConvLSTMs are inherently limited to model large, unknown transformations since we use regular CNNs with a fixed receptive field at the core of them. Hence, so as to improve the robustness of traditional ConvLSTMs to these transformations, we used deformable 2D convolutions in the core of ConvLSTMs. Additionally, since deformable convolutions require a large number of parameters, we could not use them instead of every normal ConvLSTM layer. Consequently, we decided to use deformable ConvLSTMs on certain frames only. Since the characters usually start to represent the emotion after some frames and salient parts for classification usually happen in the middle parts of a video, we decided to choose frames after 25$\%$ of video have passed, after half of the video has passed and after 75$\%$ of video have passed. In each of these parts, we chose three different frames. Moreover, as stated in \cite{12}, we removed the convolutional structure of all gates except for the input-to-state gate for spatio-temporal feature fusion which we used deformable 2D convolutions.

\subsubsection{2D CNN component}
\begin{figure}[h]
\subfloat[3D CNN Component]{
\begin{minipage}{
0.3\textwidth}
\includegraphics[clip,trim=0.72cm 20.4cm 0cm 0.65cm width=1.0\linewidth]{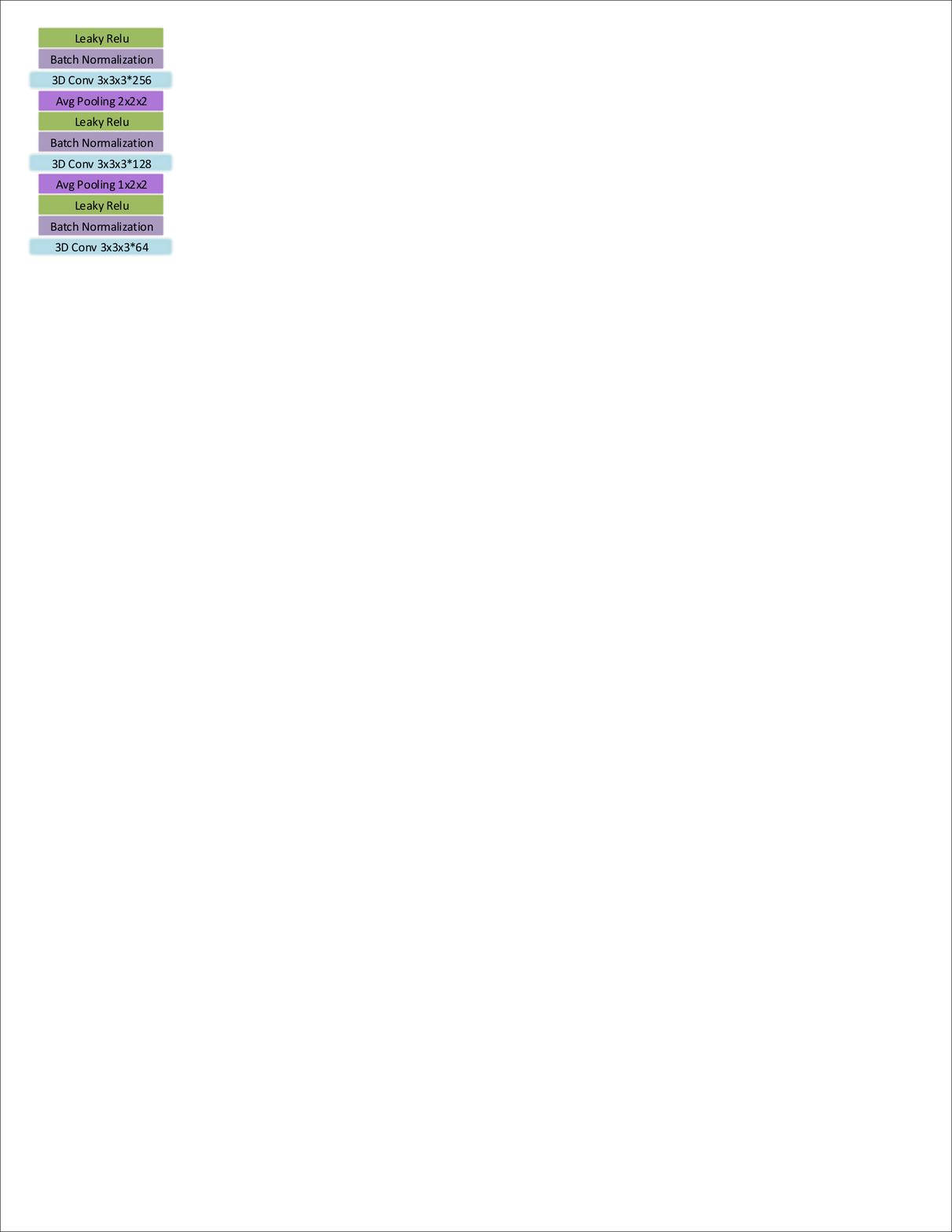}
\label{fig2a}
\end{minipage}}\hspace{5cm}
\subfloat[2D CNN Component]{\begin{minipage}{
0.3\textwidth}
\includegraphics[clip,trim=0.72cm 20.4cm 0cm 0.65cm width=1.0\linewidth]{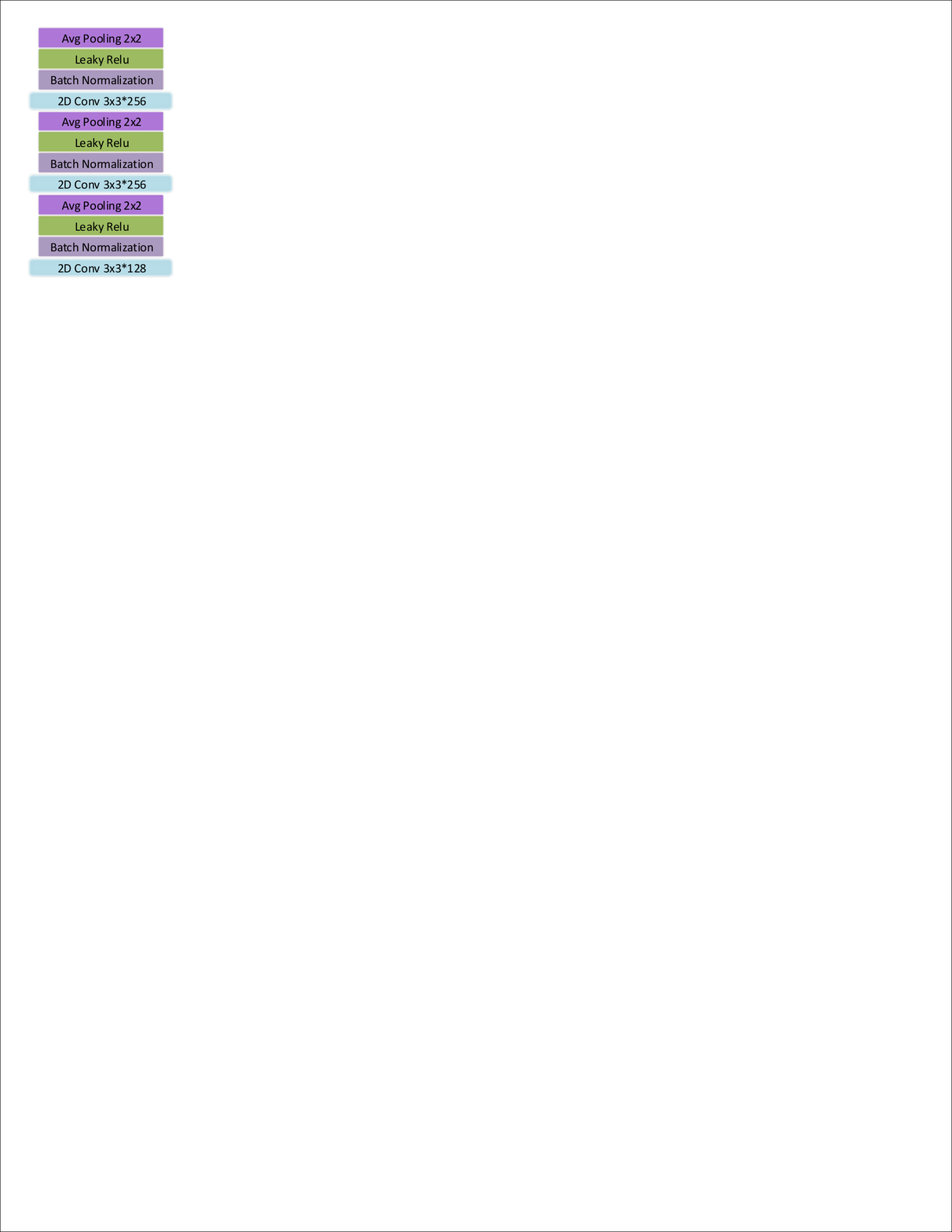}
\label{fig2b}
\end{minipage}}

\caption{(a) 3D CNN component that is used to extract short-term spatio-temporal features. (b) 2D CNN component which acts as the final classification layer.}
\label{fig2}
\end{figure}

So far, we have reduced the spatial dimension of our input to $28 \times 28$, but already, we have learned short-term and long-term spatio-temporal features. Therefore, in this part, we try to learn and focus on spatial features only using a 2D CNN component. As can be seen in Figure \ref{fig2b}, we have chosen a shallow network with three convolutional and average pooling layers. After this stage, we did a final global average pooling for final feature fusion among all frames before using a fully connected layer for final classification.

\section{Experiments}
\subsection{Dataset}
\begin{figure}[h]
\subfloat{
\begin{minipage}{
0.23\textwidth}
\includegraphics[width=1\textwidth]{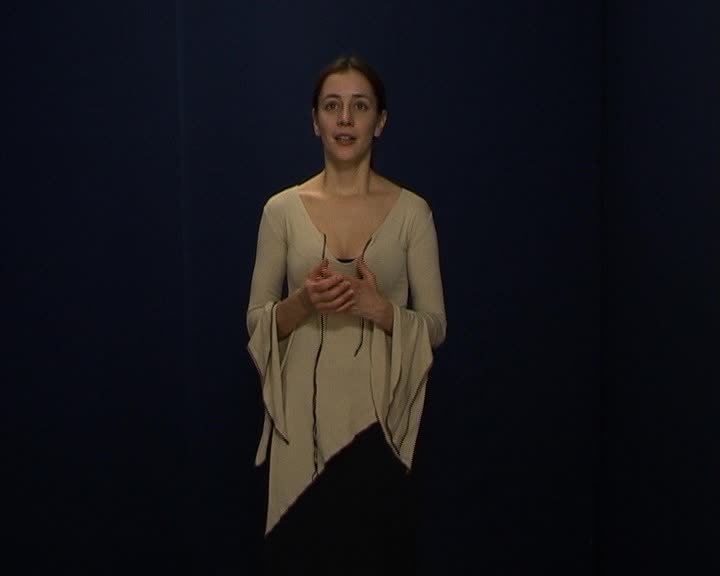}
\end{minipage}}\
\subfloat{\begin{minipage}{
0.23\textwidth}
\includegraphics[width=1\textwidth]{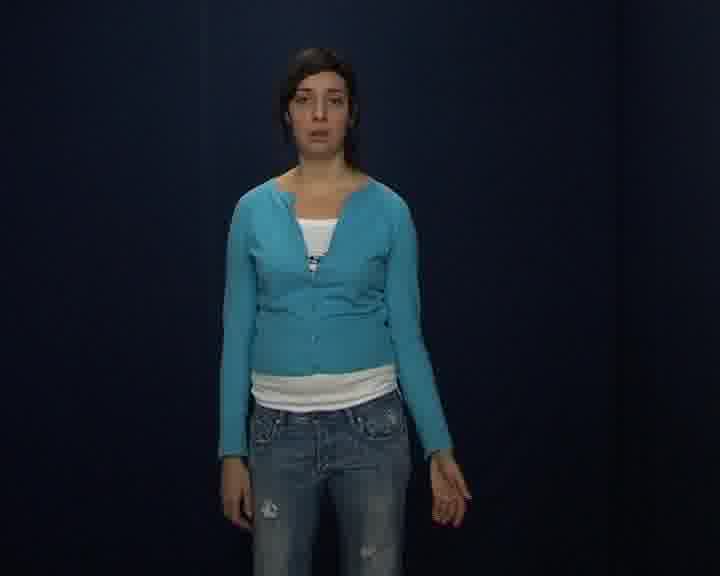}
\end{minipage}}
\subfloat{
\begin{minipage}{
0.23\textwidth}
\includegraphics[width=1\textwidth]{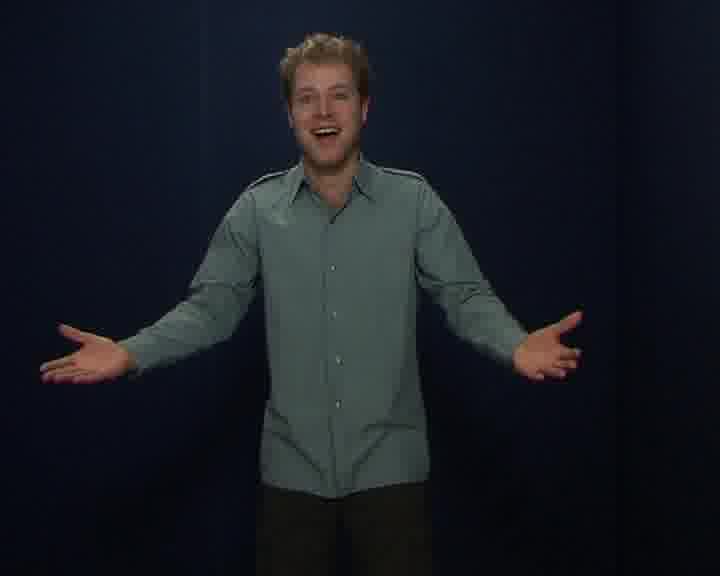}
\end{minipage}}\
\subfloat{\begin{minipage}{
0.23\textwidth}
\includegraphics[width=1\textwidth]{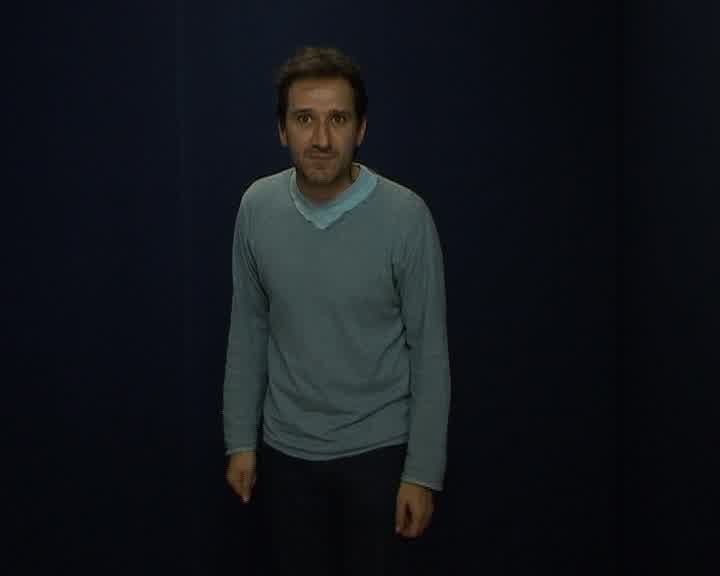}
\end{minipage}}
\caption{Some examples of the utilized GEMEP dataset in our experiments.}
\label{fig3}
\end{figure}

For experimental studies, we used the GEMEP dataset which contains 145 videos representing 17 different emotions which are called \emph{compound emotions} in the literature \cite{18}. The different emotion classes are admiration, amusement, tenderness, anger, disgust, despair, pride, anxiety, interest, irritation, joy, contempt, fear, pleasure, relief, surprise and sadness. Since the number of videos for training our network was insufficient for training a deep neural network, we did data augmentation. For augmentation, we translated each frame to four different corners by 25 pixels, rotated each video between $-30 ^\circ $ to $30 ^\circ$, used gaussian filter with different intensities and changed the brightness. Finally, we had 9052 videos which since there were not any standard division into training and validation set for this dataset, we decided to use $80\%$ for training and $20\%$ for validation. Some frames of the dataset utilized in this work are depicted in Figure \ref{fig3}.

\begin{table}[h]
\centering
\begin{tabular}{l|l}
& Accuracy \\ \hline
Normal ConvLSTM & 96.38 \\ \hline
Deformable ConvLSTM & \textbf{98.8}
\end{tabular}
\caption{Comparison between using deformable ConvLSTM and normal ConvLSTM. Note that we only applied deformable ConvLSTM to 9 different frames.}
\label{tb1}
\end{table}
\subsection{Comparison between deformable ConvLSTM and ConvLSTM}
In this section, we compare the result of using normal ConvLSTM and deformable ConvLSTM. Since deformation may happen in any given image, it makes it difficult for a fixed size receptive field to focus on the salient parts of the image. On the other hand, deformable convolutions due to their flexible nature can easily focus on these deformed parts and extract features using their arbitrary shaped receptive field. As can be seen in Table \ref{tb1}, even though we used deformable ConvLSTM only on a limited number of frames (9 frames) we were able to achieve a higher classification accuracy in comparison to normal ConvLSTM which prove their effectiveness.

\subsection{Comparison between other methods}
As can be seen in Table \ref{tb2}, our method based on deformable ConvLSTMs outperformed other methods in terms of accuracy. For comparison, we selected methods that used the whole human body for classification. The method in \cite{7} used a deep architecture for classification using images extracted from videos. The superior result of our approach indicates that in some frames the emotion may become ambiguous and by using spatio-temporal features we can clarify this ambiguity and achieve higher accuracy. Additionally, both MBMIC \cite{15} and HOG-KLT \cite{19} used frame difference to extract temporal features. However, results show that using 3D convolutions and ConvLSTMs for learning and extracting spatio-temporal features is the superior choice.

\begin{table}[h]
\centering
\begin{tabular}{l|l}
& Accuracy \\ \hline
Ours & \textbf{98.8} \\ \hline
FDCNN (Image) \cite{7} & 95.4 \\ \hline
MBMIC \cite{15} & 94.6 \\ \hline
HOG-KLT \cite{19} & 95.9
\end{tabular}
\caption{Comparison between our proposed method and other methods on the GEMEP dataset. Note that we only chose approaches which considered whole human body and the results of our method is based on validation set.}
\label{tb2}
\end{table}

\section{Conclusion}
In this paper, we proposed a method for the classification of human whole-body emotion from videos. In the first layer, we used 3D convolutions to extract short-term spatio-temporal features. Afterward, motivated by the recent success of deformable 2D convolutions we incorporated these flexible convolutions in the core of normal ConvLSTMs to form deformable ConvLSTMs. Because the receptive in these forms of convolutions are adjustable, it gives the network more freedom to extract features from different regions of a given frame. Finally, we used a shallow 2D convolutional network to further extract features from the 2D features map and used a final feature fusion. Results using GEMEP dataset show a state-of-the-art result of 98.8$\%$ accuracy on the validation set.

\par

In future works, we will try to study the effectiveness of our deformable ConvLSTMs on other classification tasks which involve videos such as human gesture recognition or facial emotion recognition. Additionally, we will explore with different frame selection strategies for using deformable ConvLSTMs so as to find the optimal choice.

%
%
%
%
%
\bibliographystyle{splncs04}
\bibliography{Refrences}

\begin{thebibliography}{10}
\providecommand{\url}[1]{\texttt{#1}}
\providecommand{\urlprefix}{URL }
\providecommand{\doi}[1]{https://doi.org/#1}

\bibitem{14}
Ahmed, F., Bari, A., Gavrilova, M.: Emotion recognition from body movement.
  IEEE Access  \textbf{PP}, ~1--1 (12 2019). \doi{10.1109/ACCESS.2019.2963113}

\bibitem{4}
Bänziger, T., Scherer, K.: Introducing the geneva multimodal emotion portrayal
  (gemep) corpus. Blueprint for Affective Computing: A Sourcebook  (01 2010)

\bibitem{9}
Chen, J., Chen, Z., Chi, Z., Fu, H.: Facial expression recognition in video
  with multiple feature fusion. IEEE Transactions on Affective Computing
  \textbf{PP}, ~1--1 (07 2016). \doi{10.1109/TAFFC.2016.2593719}

\bibitem{10}
Dai, J., Qi, H., Xiong, Y., Li, Y., Zhang, G., Hu, H., Wei, Y.: Deformable
  convolutional networks (2017)

\bibitem{18}
Du, S., Martinez, A.: Compound facial expressions of emotion: From basic
  research to clinical applications. Dialogues in Clinical Neuroscience
  \textbf{17},  443--455 (12 2015)

\bibitem{2}
Glowinski, D., Dael, N., Camurri, A., Volpe, G., Mortillaro, M., Scherer, K.:
  Toward a minimal representation of affective gestures. T. Affective Computing
   \textbf{2},  106--118 (04 2011). \doi{10.1109/T-AFFC.2011.7}

\bibitem{6}
He, K., Zhang, X., Ren, S., Sun, J.: Deep residual learning for image
  recognition (2015)

\bibitem{5}
Jain, D., Shamsolmoali, P., Sehdev, P.: Extended deep neural network for facial
  emotion recognition. Pattern Recognition Letters  \textbf{120} (04 2019).
  \doi{10.1016/j.patrec.2019.01.008}

\bibitem{8}
Jeong, D., Kim, B.G., Dong, S.Y.: Deep joint spatiotemporal network (djstn) for
  efficient facial expression recognition. Sensors  \textbf{20}(7) (2020).
  \doi{10.3390/s20071936}, \url{https://www.mdpi.com/1424-8220/20/7/1936}

\bibitem{7}
Rajaram, S., Geetha, M.: Deep learning approach for emotion recognition from
  human body movements with feedforward deep convolution neural networks.
  Procedia Computer Science  \textbf{152},  158--165 (01 2019).
  \doi{10.1016/j.procs.2019.05.038}

\bibitem{19}
Rajaram, S., Geetha, M.: Vision-Based Human Emotion Recognition Using HOG-KLT
  Feature, pp. 261--272 (01 2020). \doi{10.1007/978-981-15-3369-3-20}

\bibitem{1}
Rajaram, S., Geetha, M., J, A.: Svm-knn based emotion recognition of human in
  video using hog feature and klt tracking algorithm. International Journal of
  Pure and Applied Mathematics  \textbf{117},  621--634 (01 2017)

\bibitem{15}
Santhoshkumar, R., Geetha, M.K.: Emotion Recognition on Multi View Static
  Action Videos Using Multi Blocks Maximum Intensity Code (MBMIC), pp.
  1143--1151. Springer International Publishing, Cham (2020).
  \doi{10.1007/978-3-030-41862-5-116},
  \url{https://doi.org/10.1007/978-3-030-41862-5-116}

\bibitem{16}
Sharma, G., Dhall, A.: A Survey on Automatic Multimodal Emotion Recognition in
  the Wild, pp. 35--64. Springer International Publishing, Cham (2021).
  \doi{10.1007/978-3-030-51870-7-3},
  \url{https://doi.org/10.1007/978-3-030-51870-7-3}

\bibitem{13}
Shi, X., Chen, Z., Wang, H., Yeung, D.Y., kin Wong, W., chun Woo, W.:
  Convolutional lstm network: A machine learning approach for precipitation
  nowcasting (2015)

\bibitem{12}
Zhang, L., Zhu, G., Shen, P., Song, J.: Learning spatiotemporal features using
  3dcnn and convolutional lstm for gesture recognition. pp. 3120--3128 (10
  2017). \doi{10.1109/ICCVW.2017.369}

\bibitem{11}
Zhu, X., Hu, H., Lin, S., Dai, J.: Deformable convnets v2: More deformable,
  better results (2018)

\end{thebibliography}

\end{document}